\documentclass[sigconf,natbib=true,anonymous=false]{acmart}

\makeatletter
\def\@ACM@checkaffil{
    \if@ACM@instpresent\else
    \ClassWarningNoLine{\@classname}{No institution present for an affiliation}%
    \fi
    \if@ACM@citypresent\else
    \ClassWarningNoLine{\@classname}{No city present for an affiliation}%
    \fi
    \if@ACM@countrypresent\else
        \ClassWarningNoLine{\@classname}{No country present for an affiliation}%
    \fi
}
\makeatother

\def\CHK#1 {\textcolor{magenta}{{\bf [CHK:}~#1{\bf ]}}~}
\def\ADD#1 {\textcolor{cyan}{{\bf [ADD:}~#1{\bf}]}~}

\usepackage[utf8]{inputenc}

\usepackage{subcaption} 

\usepackage{multirow} 

\usepackage{amsmath,empheq} 

\usepackage[ruled,linesnumbered]{algorithm2e} 
\SetKwComment{Comment}{/* }{ */} 

\usepackage[title]{appendix}

\usepackage{enumitem}
\setlist[itemize]{leftmargin=*}
\setlist[enumerate]{leftmargin=*}

\renewcommand{\figureautorefname}{Figure}

\def\equationautorefname~#1\null{Eqn. ~(#1)\null}
\def\figureautorefname~#1\null{Fig. ~#1\null}

\newtheoremstyle{remboldstyle}
  {}{}{\itshape}{}{\bfseries}{:}{.5em}{{\thmname{#1 }}{\thmnumber{#2.}}{\thmnote{ #3}}}
\theoremstyle{remboldstyle}

\copyrightyear{2023} 
\acmYear{2023} 
\acmConference[MM '23]{Proceedings of the 31st ACM International Conference on Multimedia}{October 29-November 3, 2023}{Ottawa, ON, Canada}
\acmBooktitle{Proceedings of the 31st ACM International Conference on Multimedia (MM '23), October 29-November 3, 2023, Ottawa, ON, Canada}
\acmPrice{15.00}

\begin{document}

\begingroup
\hyphenpenalty 9000
\exhyphenpenalty 9000

\title{Temporal Sentence Grounding in Streaming Videos}

\author{Tian Gan}
\email{gantian@sdu.edu.cn}
\affiliation{%
  \institution{Shandong University}
}

\author{Xiao Wang}
\email{scz.wangxiao@gmail.com}
\affiliation{%
  \institution{Harbin Institute of Technology, Shenzhen}
}

\author{Yan Sun}
\email{sy1471371562@gmail.com}
\affiliation{%
  \institution{Shandong University}
}

\author{Jianlong Wu}
\email{jlwu1992@pku.edu.cn}
\authornote{Corresponding author: Jianlong Wu.}
\affiliation{%
  \institution{Harbin Institute of Technology, Shenzhen}
}

\author{Qingpei Guo}
\email{qingpei.gqp@antfin.com}
\affiliation{%
  \institution{Ant Group}
}

\author{Liqiang Nie}
\email{nieliqiang@gmail.com}
\affiliation{%
  \institution{Harbin Institute of Technology, Shenzhen}
}

\renewcommand{\shortauthors}{Tian Gan et al.}

\begin{abstract}
This paper aims to tackle a novel task - Temporal Sentence Grounding in Streaming Videos (TSGSV). 
The goal of TSGSV is to evaluate the relevance between a video stream and a given sentence query. 
Unlike regular videos, 
streaming videos are acquired continuously from a particular source, and are always desired to be processed on-the-fly in many applications such as surveillance and live-stream analysis. 
Thus, TSGSV is challenging since it requires the model to infer without future frames and process long historical frames effectively, 
which is untouched in the early methods.
To specifically address the above challenges,
we propose two novel methods:
(1) a TwinNet structure that enables the model to learn about upcoming events; 
and
(2) a language-guided feature compressor that eliminates redundant visual frames and reinforces the frames that are relevant to the query.
We conduct extensive experiments using ActivityNet Captions, TACoS, and MAD datasets. The results demonstrate the superiority of our proposed methods.
A systematic ablation study also confirms their effectiveness.
\end{abstract}



%

\begin{abstract}
This paper aims to tackle a novel task - Temporal Sentence Grounding in Streaming Videos (TSGSV). 
The goal of TSGSV is to evaluate the relevance between a video stream and a given sentence query. 
Unlike regular videos, 
streaming videos are acquired continuously from a particular source, and are always desired to be processed on-the-fly in many applications such as surveillance and live-stream analysis. 
Thus, TSGSV is challenging since it requires the model to infer without future frames and process long historical frames effectively, 
which is untouched in the early methods.
To specifically address the above challenges,
we propose two novel methods:
(1) a TwinNet structure that enables the model to learn about upcoming events; 
and
(2) a language-guided feature compressor that eliminates redundant visual frames and reinforces the frames that are relevant to the query.
We conduct extensive experiments using ActivityNet Captions, TACoS, and MAD datasets. The results demonstrate the superiority of our proposed methods.
A systematic ablation study also confirms their effectiveness.
\end{abstract}

\maketitle

\section{Introduction} \label{sec:introduction}

In the past few decades, streaming videos have experienced significant growth, with examples including surveillance and live-stream videos. 
%
Nowadays, 
it is estimated that there are more than one billion surveillance cameras operating worldwide, while the second quarter of 2022 saw people collectively spending more than 5 billion hours on live streaming video platforms\footnote{Industry research reports from \href{https://shorturl.at/sTUV2}{visualcapitalist}, \href{https://www.statista.com/statistics/1284059/usa-live-video-viewership/}{statista}, and \href{https://www.askci.com/news/chanye/20220318/1416321746317.shtml}{askci}.}.
In this context, the development of algorithms capable of detecting the exact moment a queried event occurs in real-time video streams has become paramount. 
Temporal Sentence Grounding in Streaming Videos (TSGSV) is one such algorithm that aims to address this challenge.
Compared to conventional videos, streaming videos are continuously sourced and can run infinitely. 
Given their critical importance in many practical applications, it is necessary to have online models that can process them on-the-fly for runtime monitoring~\cite{shou_sample_2018,Wang_TMM_2022,Gan_tomccap_19,Dong_CVPR_SHA}.

\begin{figure}[t]
    \centering
    \includegraphics[width=0.9\linewidth]{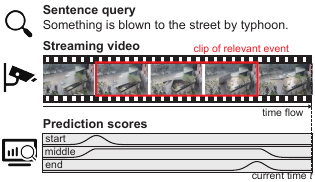}
    \vspace{-1em}
    \caption{An example of the task of Temporal Sentence Grounding in Streaming Videos (TSGSV).
    The TSGSV model is to estimate immediately the probability of the current time being the start, middle, or end of an relevant event.
    }
    \label{fig:task_example}
\end{figure}


It is worth mentioning that there has been a surge of research in relevant fields, 
including streaming video understanding and temporal sentence grounding (offline version of TSGSV). 
However, their models cannot be applied directly to our task.
Regarding \textit{streaming video understanding}, 
existing research efforts have been concentrated either on the recognition of a limited set of pre-defined actions~\cite{geest_OAD_2016, shou_sample_2018, xu_lstr_2021, wu2023neighbor}, 
or on the extraction of salient clips (highlights) from the video~\cite{zhao_antpivot_2022}. 
However, these models suffer from the limitation of their failure to generate responses to arbitrary sentence queries. 
In the context of \textit{temporal sentence grounding}, 
existing methods necessitate the availability of the entire video in advance, thereby precluding their capability to support online inference.

In this paper, we formulate the task of temporal sentence grounding in streaming videos,
which evaluates the relevance between a given query and the video stream in an online fashion.
%
For example, given a query ``\textit{Something is blown to the street by typhoon.}'', TSGSV model estimates immediately the probability of current time being the start, middle, or end of the relevant event, as illustrated in \autoref{fig:task_example}. 

TSGSV is, however, non-trivial due to the following challenges: 
\textbf{1) Incomplete Information Inference}. 
In the streaming video scenario, 
a model has to make predictions based on incomplete information available up until that moment.
Consequently, 
the model cannot leverage information from future frames, 
rendering the task significantly more challenging than its offline counterpart.
Pioneer studies on online action detection \cite{xu_TRN_2019, wang_oadtr_2021} 
approach this same challenge 
by simultaneously performing \textit{future prediction} (of action label). 
Drawing inspiration from human cognitive behavior, their hypothesis is that 
``\textit{explicitly forecasting the future can facilitate more accurate action classification in the present}"~\cite{xu_TRN_2019}. 
%
However, 
as we have verified in our experiments, simply conducting future prediction as they did actually harms the grounding performance in TSGSV.
Considering that the query sentence would have arbitrary words instead of a closed action set, 
future prediction would be much more difficult thus the previous methods are not robust in this scenario.
Thereby, how to devise a model that can perceive future events while improving grounding performance is the first challenge. 
\textbf{2) Historical Information Compression}. 
Historical frames can provide valuable contextual information for current events. 
However, streaming videos generally have extremely long historical frames, most of which are irrelevant to the current events. Accordingly, 
to decrease computational complexity and increase information density, we can compress them by shortening their length. 
Existing compression method for streaming video \cite{xu_lstr_2021} has not take sentence query into consideration, 
resulting in query-irrelevant information. 
Therefore, how to effectively compress historical information for better representation is the second challenge.

To this end, we design a TwinNet architecture and a Language-guided Feature Compressor (LFC) to tackle the above two challenges, respectively. 
The TwinNet architecture is designed to handle the recognition of arbitrary sentence queries while mitigating the requirement for complete video availability.
In particular, the TwinNet architecture contains 
an ordinary network and a prophet network.
The former takes video frames up to the present time as input 
while the latter additionally utilizes future frames. 
Through knowledge distillation during training, the prophet network teaches the ordinary network to anticipate future events, though it is removed during the inference phase. 
%
Furthermore, the LFC component compresses historical frames under the guidance of a sentence query. 
Specifically, the LFC first employs two parallel modules to respectively suppress visually redundant information and strengthen query-relevant frames. 
Subsequently, LFC combines their outcomes with dynamic routers by considering the significance of both modules.

In a nutshell, we summarize our threefold contributions:
\begin{itemize}
    \item 
     We formulate the challenging task of \textit{Temporal Sentence Grounding in Streaming Videos}, thus paving the way for future research in the field. 
     Our work marks the \textbf{first} attempt to address this problem and offers promising opportunities for various streaming video applications.
     
    \item 
    We propose a novel TwinNet architecture and a language-guided feature compressor that address key challenges in TSGSV. The TwinNet architecture is specifically designed to tackle the problem of incomplete information inference, while the language-guided feature compressor successfully compresses historical information, thus enabling faster and more efficient processing of video data.
    
    \item Extensive experiments on ActivityNet Captions, TACoS, and MAD datasets demonstrate the superiority of our proposed methods. 
    Moreover, we make our code openly available to facilitate reproducibility and further research efforts\footnote{See our GitHub repository \href{https://github.com/SCZwangxiao/TSGVs-MM2023}{https://github.com/SCZwangxiao/TSGVs-MM2023}.}.        
\end{itemize}


    
\section{Related Work}
\subsection{Temporal Sentence Grounding in Videos}
The task of Temporal Sentence Grounding in Videos (TSGV) aims to localize a clip in a video which corresponds to a given sentence query. Pioneer work can be broadly classified into three categories based on the clip proposal generation strategy \cite{zhang_aixin_review_2022, yuan_survey2_2021, lan_survey3_2022}: proposal-based, proposal-free, and reinforcement learning methods.

Prior research in TSGV has explored proposal-based methods in which candidate clips are first generated from sliding windows~\cite{gao_ctrl_2017, hendricks_mcn_2017, liu_role_2018} or proposal network~\cite{xu_qspn_2019},
and subsequently ranked based on a given query. 
Recent efforts have moved towards embedding proposal generation within ranking modules for end-to-end learning~\cite{chen_tgn_2018, zhang_2dtan_2020, zhang_ms2dtan_2021, wang_dpin_2020}.
However, proposal-based methods are not suitable for streaming videos as they require a complete video for candidate clip proposal generation, 
and therefore cannot be applied in scenarios where video is being simultaneously recorded and processed.

Proposal-free methods have two variants: regression-based and span-based. 
Regression-based methods \cite{yuan_ablr_2019, qu_fian_2020, zhang_new1_2023, pan_new2_2023, luo_new3_2023} directly predict the distance between each timestamp and the ground-truth clip. 
However, these methods face a challenge in the context of streaming videos where the ground-truth clip may occur in unseen future frames, rendering such a regression target unrealistic. 
%
In contrast,
span-based methods \cite{ghosh_excl_2019, chen_lnet_2019, zhang_vslnet_2020, zhang_vslnetL_2021, zhao_CPN_2021} aim to predict the probability of each timestamp representing either the start or end point of the ground-truth clip. 
While the prediction target for span-based methods does not rely on future frames, these approaches necessitate access to the entire video for contextual information.

Reinforcement learning techniques, as demonstrated in several studies \cite{he_rwmrl_2019, wang_smrl_2019, cao_strong_2020, wu_bar_2020}, aim to replicate the decision-making strategies employed by humans.
Specifically, TSGV is treated as a sequential decision-making problem, where the model adjusts the position of candidate clips along the video based on their relevance to the query. 
Although this formulation appears to be compatible with streaming data, two main challenges currently hinder its practical application. 
First, these methods require access to the complete video to perform coarse-to-fine clip refinement. 
Second, existing RL methods necessitate a meticulous design of the action space \cite{wu_tsp_prl_2020} or reward function \cite{sun_maban_2021} to achieve better performance, which is difficult to optimize. 
Therefore, these approaches may not be feasible for our specific TSGV problem.

\subsection{Video Understanding for Streaming Videos}
Comprehending streaming video content entails a different approach than traditional video understanding, as the model must analyze the video without access to future frames. To the best of our knowledge, there exist primarily three video understanding tasks for streaming videos, namely online action detection~\cite{geest_OAD_2016}, online temporal action localization~\cite{shou_sample_2018}, and livestream highlight detection~\cite{zhao_antpivot_2022}.


The online action detection task, as defined in \cite{geest_OAD_2016}, predicts the class labels for video frame real-time. One of the primary challenges inherent to this task is the lack of access to future frames, motivating some methods \cite{xu_TRN_2019, wang_oadtr_2021} to incorporate anticipated future information to enhance prediction accuracy. The work PKD \cite{ppk} inserts privileged knowledge to relax the distillation loss function. They assume that the same output for both offline and online models will result in over-fitting.
To address the challenge of similar backgrounds between consecutive actions, various methods have been proposed. 
For example, Shou \textit{et al.} \cite{shou_sample_2018} introduced a GAN-based method to generate hard-negative samples for training. 
Eun \textit{et al.} \cite{eun_IDN_2020} proposed an information discrimination unit to accumulate relevant information specifically for the current action. 
Chen \textit{et al.} \cite{chen_gatehub_2022} designed a gated history unit to suppress background information. 
By contrast, Xu \textit{et al.} \cite{xu_lstr_2021} utilized a long-short-term transformer to compress the most valuable historical information while discarding the rest. We note that the online action detection task is often restricted to a closed action set, rendering it unsuitable for handling arbitrary sentence queries.


The concept of online temporal action localization has been introduced in a recent study by Kang~\textit{et al.}~\cite{kang_cag-qil_2021}. The main goal of this task is to produce action instances in conjunction with their corresponding start and end timestamps. 
This process adopts the framework of Markov decision process and is optimized using Q-imitation learning. Despite its ability to extract video clips, this particular approach is limited to a pre-defined closed action set.
One potential drawback of confining action localization to a pre-established set is the inability to detect newly emerging actions in real-time. This can pose a significant challenge, particularly in dynamic situations, where a flexible framework capable of detecting unforeseen actions is required. As such, the development of an approach that enables the identification of novel actions in real-time would be an important contribution.


Livestream highlight detection is proposed by Yang \textit{et al.} \cite{zhao_antpivot_2022}. The goal of this task is to automatically extract salient or valuable video clips. To accomplish this, they formulated the task as a two-step process of segmentation and importance evaluation along the temporal axis of the livestream. The authors proposed a novel architecture, named AntPivot, which utilizes a hierarchical attention module to address the highlight detection task in a dynamic-programming fashion. Although the approach is capable of producing clips from an open set of videos, it does not incorporate an understanding of the semantic content within the detected clips.

\section{Methodology} \label{sec:method}

\subsection{Problem Formulation}
Given a streaming video and a sentence query, our goal is to evaluate the relevance of them.
Formally, a streaming video at time $T$ is represented by a sequence of $\tau$ past frames, 
$\mathcal{I}_T=\{I_{T-\tau+1}, ...,I_T\}$, 
and a sentence query is represented by a sequence of $N$ tokens,
$\mathcal{Q}=\{w_1, ..., w_N\}$. 
The temporal sentence grounding system will estimate $s_T,m_T,e_T\in\mathbb{R}$, 
i.e., 
the probability of time $T$ being the start, middle, and end point in the streaming video, respectively.

\subsection{Overview}

The overall architecture is illustrated in \autoref{fig:overview}(a). 
Our model comprises twin networks, namely the ordinary and prophet networks. 
Each network is composed of a feature compressor, decoder, and predictor.
The TwinNet architecture is designed to tackle the challenge of \textit{incomplete information inference}.
In the training phase, 
the prophet network is enabled to access the future frames and subsequently teach the ordinary network in comprehending future events.
During the inference phase, the prophet network will be disregarded.
To meet the requirement of \textit{historical information compression} challenge,
we design a language-guided feature compressor that is applied in both twin networks.
In the following sections, we will detail TwinNet and its constituent components.

\subsection{Preprocessing}
For language features, we apply a pre-trained language model to encode each sentence token for word features $\mathbf{Q}\in\mathbb{R}^{N\times d}$, where $d$ is embedding dimensions. 
Subsequently, a LSTM is utilized to generate the \textit{query feature},
denoted by $\mathbf{q} \in \mathbb{R}^d$:
\begin{equation}
    \mathbf{q} = \textmd{LSTM}(\mathbf{Q}).
\end{equation}

For video features, we apply a pre-trained feature extractor to process each frame in $\mathcal{I}_T$ into a feature vector for frame-level features $\mathbf{V}\in\mathbb{R}^{\tau\times d}$:
\begin{equation}
    \mathbf{V}=[\mathbf{f}_{T-\tau+1}, ...,\mathbf{f}_T].
\end{equation}

Then, 
following the online action detection method proposed in~\cite{xu_lstr_2021},
we partition $\mathbf{V}$ into two subsets: the \textit{present feature} denoted as $\mathbf{V}_p\in\mathbb{R}^{M_p\times d}$ and the \textit{historical feature} denoted as $\mathbf{V}_h\in\mathbb{R}^{M_h\times d}$:
\begin{empheq}[left=\empheqlbrace]{align}
    & \mathbf{V}_p = \left [  \mathbf{f}_{T-M_p+1}, ..., \mathbf{f}_{T}  \right ], \\
    & \mathbf{V}_h = \left [  \mathbf{f}_{T-M_p-M_h+1}, ..., \mathbf{f}_{T-M_p}  \right ],
\end{empheq}
where $M_p<M_h$ 
denote the lengths of the present and historical features, respectively. 
Intuitively, the present feature focuses on the current events, while the historical feature carries contextual information.

For prophet network, we additionally incorporate \textit{future feature} $\mathbf{V}_f\in\mathbb{R}^{M_h\times d}$:
\begin{equation}
    \mathbf{V}_f = \left [  \mathbf{f}_{T+1}, ..., \mathbf{f}_{T+M_h} \right ].
\end{equation}

Note that we add linear transformation layers with GeLU activation 
after text and vision feature extractor to map video feature
$\mathbf{V}$ and 
language feature $\mathbf{Q}$
into a common $d$-D embedding space.

\subsection{TwinNet Architecture}

\begin{figure*}[t]
    \centering
    \includegraphics[width=0.99\linewidth]{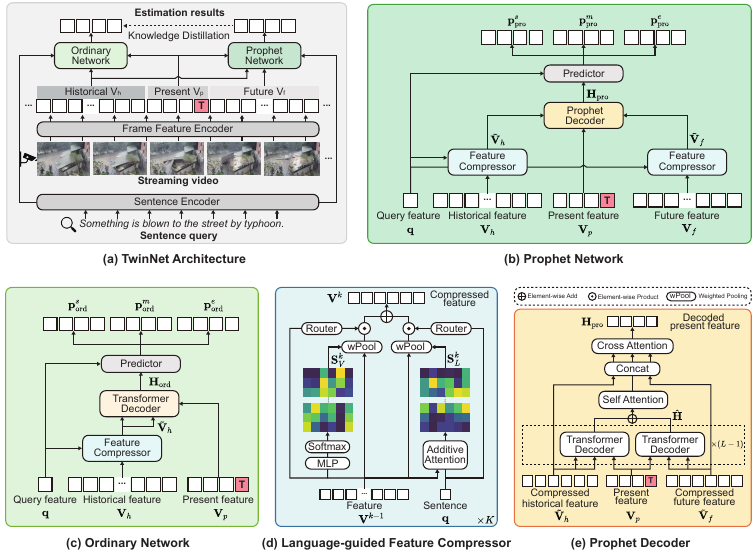}
    \caption{
    The overview of our method. 
    It has a \textbf{TwinNet architecture}(a) incorporates \textbf{ordinary network}(c) and \textbf{prophet network}(b). Both networks have \textbf{language-guided feature compressor}(d),
    whereas  the \textbf{prophet decoder}(e) is designed for the prophet network.
    }
    \label{fig:overview}
\end{figure*}

\subsubsection{Ordinary Network}
As illustrated in \autoref{fig:overview} (c), 
the ordinary network is designed to estimate the relevance between the sentence query and what is happening at present time $T$. 
Therefore, we need to combine current information in the present feature
$\mathbf{V}_p$ 
and contextual information from the historical feature 
$\mathbf{V}_h$
for a holistic understanding. 
Besides, to increase information density in $\mathbf{V}_h$ and reduce computational complexity, we need to compress $\mathbf{V}_h$ 
for better efficiency.

Concretely, we first derive the compressed historical feature $\tilde{\mathbf{V}}_h\in\mathbb{R}^{n\times d}$ under the guidance of query $\mathbf{q}$:
\begin{equation} 
    \tilde{\mathbf{V}}_h = \textmd{FeatureCompressor} \left ( \mathbf{V}_h, \mathbf{q} \right ),
\end{equation}
where $n < M_h$ is the length of the compressed feature.
Afterwards, we use $\mathbf{V}_p$ as a query to decode relevant information from $\tilde{\mathbf{V}}_h$, obtaining the holistic present feature in ordinary network $\mathbf{H}_{\textmd{ord}}\in\mathbb{R}^{M_p\times d}$:
\begin{equation} \label{eq:ord_decoder}
    \mathbf{H}_{\textmd{ord}} = \textmd{TransformerDecoder} \left ( \mathbf{V}_p, \tilde{\mathbf{V}}_h, \tilde{\mathbf{V}}_h \right ).
\end{equation}
The TransformerDecoder has $L$ standard transformer decoder layers. Note that we apply an attention mask to $\mathbf{V}_p$ so that any frame in the present feature can only depend on its previous frames.

Finally, in the predictor component, we can estimate the probability of present time being start $\textbf{p}^s_{\textmd{ord}}\in\mathbb{R}^{M_p}$, middle $\textbf{p}^m_{\textmd{ord}}\in\mathbb{R}^{M_p}$, and end $\textbf{p}^e_{\textmd{ord}}\in\mathbb{R}^{M_p}$ using product similarity:
\begin{equation} \label{eq:ord_predictor}
    \textbf{p}^{\xi}_{\textmd{ord}} = \textbf{H}^{ }_{\textmd{ord}} \textbf{W}^{\xi}_{\textmd{ord}} \textbf{q},\ \xi\in\{s,e,m\},
\end{equation}
where $\textbf{W}^{\xi}_{\textmd{ord}}\in\mathbb{R}^{d\times d}$ is the transformation matrix.

It should be noted that in~\autoref{eq:ord_predictor}, the probabilities for every frame are estimated from present frames rather than solely from the latest frame at time $T$.
This setting enables the model to utilize more supervision signals during the training phase.
During inference, however, only the probabilities for frame at time $T$ are computed.




\subsubsection{Prophet Network}
As illustrated in \autoref{fig:overview} (b), 
the prophet network shares a similar pipeline with the ordinary network,
with the exception that it can access future frames.

To be more specific, 
we start by deriving the compressed future feature $\tilde{\mathbf{V}}_f\in\mathbb{R}^{n\times d}$ guided by the query feature:
\begin{equation} 
    \tilde{\mathbf{V}}_f = \textmd{FeatureCompressor} \left ( \mathbf{V}_f, \mathbf{q} \right ).
\end{equation} 
Next, to combine $\mathbf{V}_p$ with contextual information from both historical and future features, we introduce the Prophet decoder to derive the holistic present feature
$\mathbf{H}_{\textmd{pro}}\in\mathbb{R}^{M_p\times d}$
in the prophet network:
\begin{equation} \label{eq:pre_dec_abs}
    \mathbf{H}_{\textmd{pro}} = \textmd{ProphetDecoder} \left ( \mathbf{V}_p, \tilde{\mathbf{V}}_h, \tilde{\mathbf{V}}_f \right ).
\end{equation}

Finally, in the predictor component, we use product similarity to estimate the probability of the present time as the start $\textbf{p}^s_{\textmd{pro}}\in\mathbb{R}^{M_p}$, middle $\textbf{p}^m_{\textmd{pro}}\in\mathbb{R}^{M_p}$, and end $\textbf{p}^e_{\textmd{pro}}\in\mathbb{R}^{M_p}$:
\begin{equation} \label{eq:predictor}
    \textbf{p}^{\xi}_{\textmd{pro}} = \textbf{H}^{ }_{\textmd{pro}} \textbf{W}^{\xi}_{\textmd{pro}} \textbf{q},\ \xi\in\{s,e,m\},
\end{equation}
where $\textbf{W}^{\xi}_{\textmd{pro}}\in\mathbb{R}^{d\times d}$ is the transformation matrix.

\subsection{Components}
\subsubsection{Language-guided Feature Compressor}
We propose a $K$-layer feature compressor. 
It aims to compress the historical/future feature 
$\mathbf{V}$
into the historical/future compressed feature 
$\Tilde{\mathbf{V}}$.
This process is accomplished by using a sentence query as a guide.
Specifically, 
we first compress the feature by suppressing visually-redundant frames 
(by the vision branch) 
and strengthening query-relevant frames 
(by the language branch). 
Lastly, we utilize a fusion network to evaluate the importance of each branch and fuse their respective results. 

The proposed architecture is illustrated in \autoref{fig:overview} (d).
We denote the output of the 
$k$-th 
layer as 
$\mathbf{V}^k\in\mathbb{R}^{n_k\times d}$, 
where 
$n_k$ 
is the length of the feature. 
In particular, 
we define 
$\mathbf{V}^0=\mathbf{V}$ 
and 
$\Tilde{\mathbf{V}} = \mathbf{V}^K$, 
such that $n_0=M_h$ and $n_k=n, k=1,...,K$.

\textbf{Vision branch.}
We adopt the TokenLearner~\cite{Michael_tokenlearner_2021} as our vision branch. 
Specifically,
we treat 
$\mathbf{V}^{k-1}$ 
and 
$\mathbf{V}^{k}$ 
as visual token sequences of length 
$n_{k-1}$ and 
$n_k$, respectively,
and compute the importance logits of each input token with respect to the output tokens 
$\textbf{E}^k_V\in\mathbb{R}^{n_{k-1}\times n_k}$ 
using a Multi-Layer Perceptron (MLP):
\begin{equation}
    \textbf{E}^k_V = \textmd{MLP}^k_V \left ( \mathbf{V}^{k-1} \right ),
\end{equation}
where the MLP is composed of two fully connected layers. 
Next, we calculate the importance score
$\textbf{S}^k_V\in\mathbb{R}^{n_{k-1}\times n_k}$ 
by applying the softmax function along input tokens:
\begin{equation}
    (\textbf{S}^k_V)_{i,j} = 
    \frac{\exp{(\textbf{E}^k_V)_{i,j}}}
    {\sum_{p=1}^{p=n_{k-1}} \exp{(\textbf{E}^k_V)_{p,j}}}.
\end{equation}
Finally, 
we compute the compressed
output of the vision branch $\mathbf{V}^k_V\in\mathbb{R}^{n_k\times d}$ 
by accumulating the input feature 
$\mathbf{V}^{k-1}$ 
according to the importance score:
\begin{equation} \label{eq:vision_branch_bmm}
    \mathbf{V}^k_V = \left ( \textbf{S}^k_V \right )^T \mathbf{V}^{k-1}.
\end{equation}

\textbf{Language branch.}
We design language branches using a similar approach with vision branch. 
To measure the relevance between 
$\mathbf{V}^{k-1}$ 
and query 
$\textbf{q}$, 
we first calculate the importance logits $\textbf{E}^k_L\in\mathbb{R}^{n_{k-1}\times n_k}$ as follows:
\begin{empheq}[left=\empheqlbrace]{align} 
    & \Tilde{\textbf{e}}^k_i = \mathbf{W}^k_1 \textbf{v}_i^{k-1} + 
    \mathbf{W}^k_2 \mathbf{q} + \mathbf{b}^k, i=1,...,n_{k-1}, \\
    & \textbf{e}^k_i = \mathbf{W}^k_3 \textmd{Tanh} \left ( \Tilde{\textbf{e}}^k_i \right ), i=1,...,n_{k-1}, \\
    & \textbf{E}^k_L = \left [ 
    \textbf{e}^k_1,
    ..., 
    \textbf{e}^k_{n_{k-1}} \right]^T,
\end{empheq}
where $\textbf{v}_i^{k-1}\in\mathbb{R}^d$ is the $i$-th token of $\mathbf{V}^{k-1}$, $\mathbf{W}^k_1\in\mathbb{R}^{d\times d}$, $\mathbf{W}^k_2\in\mathbb{R}^{d\times d}$, $\mathbf{W}^k_3\in\mathbb{R}^{n_k\times d}$ are transformation matrices, and $\mathbf{b}^k\in\mathbb{R}^d$ is the bias vector.

In a similar manner to the vision branch, 
we can derive the compressed output of the language branch,
which we denoted as
$\mathbf{V}^k_L\in\mathbb{R}^{n_k\times d}$:
\begin{empheq}[left=\empheqlbrace]{align} 
    & (\mathbf{S}^k_L)_{i,j} = 
        \frac{\exp{(\textbf{E}^k_L)_{i,j}}}
        {\sum_{p=1}^{p=n_{k-1}} \exp{(\textbf{E}^k_L)_{p,j}}}, \\
    & \mathbf{V}^k_L = \left ( \textbf{S}^k_L \right )^T \mathbf{V}^{k-1}. \label{eq:language_branch_bmm}
\end{empheq}

\textbf{Fusion Network.}
We utilize two router networks~\cite{qu_dime_2021} to dynamically adjust the importance weights of the vision branch 
$g^k_V\in\mathbb{R}$
and the language branch 
$g^k_L\in\mathbb{R}$:
\begin{empheq}[left=\empheqlbrace]{align} 
    & g^k_V = 
    \sigma \left \{
        \textmd{Tanh} \left [
            \textmd{MLP} \left ( 
            \textmd{MeanPool}\left ( \mathbf{V}^{k-1} \right )
            \right )
        \right]
    \right \}, \\
    & g^k_L = 
    \sigma \left \{
        \textmd{Tanh} \left [
            \textmd{MLP} \left ( 
            \textbf{q}
            \right )
        \right]
    \right \},
\end{empheq}
where $\sigma$ is the GeLU activation function. The final output in the $k$-th layer is calculated as:
\begin{equation}
    \mathbf{V}^k = g^k_V * \mathbf{V}^k_V + g^k_L * \mathbf{V}^k_L.
\end{equation}

\subsubsection{Prophet Decoder}
We propose a prophet decoder to combine incident information in the present feature 
$\mathbf{V}_p$ 
and contextual information from compressed historical features
$\tilde{\mathbf{V}}_h$
and future features $ \tilde{\mathbf{V}}_f$, 
to derive the holistic present feature $\mathbf{H}_{\textmd{pro}}$. 
The decoder consists of two stages: 
separate and united modeling stage, 
as shown in \autoref{fig:overview} (e). 
In the former stage, 
the present feature is combined with compressed historical and future features separately. 
In the latter stage, 
we combine the present features with united contextual features.

In the separate modeling stage, 
we apply two transformer decoders with 
$(L-1)$ 
layers to retrieve relevant contextual info from historical and future features, 
respectively:
\begin{empheq}[left=\empheqlbrace]{align} 
    & \mathbf{H}^h = \textmd{TransformerDecoder} \left ( \mathbf{V}_p, \tilde{\mathbf{V}}_h, \tilde{\mathbf{V}}_h \right ), \label{eq:pro_h_decoder} \\
    & \mathbf{H}^f = \textmd{TransformerDecoder} \left ( \mathbf{V}_p, \tilde{\mathbf{V}}_f, \tilde{\mathbf{V}}_f \right ). \label{eq:pro_f_decoder}
\end{empheq}
Note that the TransformerDecoders in \autoref{eq:ord_decoder}, \autoref{eq:pro_h_decoder}, 
and \autoref{eq:pro_f_decoder} do NOT share parameters.

In the united modeling stage, 
we merge 
$\mathbf{H}^h$ 
and 
$\mathbf{H}^f$ 
using concatenation and self-attention:
\begin{empheq}[left=\empheqlbrace]{align}
    & \hat{\mathbf{H}} = \textmd{FC} \left( \left[
    \mathbf{H}^h; \mathbf{H}^f \right] \right), \\
    & \tilde{\mathbf{H}}_{\textmd{pro}} = \textmd{SelfAttention} \left ( 
    \hat{\mathbf{H}}, \hat{\mathbf{H}}, \hat{\mathbf{H}} \right ).
\end{empheq}

Finally, we treat the present features 
$\tilde{\mathbf{H}}_{\textmd{pro}}$ 
as the query and the vector concatenation of compressed historical and future memories as key-value, 
resulting in the holistic present representation 
$\mathbf{H}_{\textmd{pro}}$
in the prophet network, as follows:
\begin{empheq}[left=\empheqlbrace]{align}
    & \tilde{\mathbf{V}} = \textmd{FC} \left( \left[
    \tilde{\mathbf{V}}_h; \tilde{\mathbf{V}}_f \right] \right), \\
    & \mathbf{H}_{\textmd{pro}} = \textmd{CrossAttention} \left ( 
    \tilde{\mathbf{H}}_{\textmd{pro}}, \tilde{\mathbf{V}}, \tilde{\mathbf{V}} \right ),
\end{empheq}




\subsection{Training and Inference} 

\subsubsection{Training}
\label{sec:train_infer}
The training sample $\{\mathcal{I}^T, \mathcal{Q}\}$ and ground-truth labels are constructed as follows. 
We first randomly sample a time point $T$ in the given video for each query sentence. 
Then, we fill the historical and present features by tracing back with $M_h+M_p$ frames, and fill the future feature by moving forward $M_h$ frames. 
However, since the ground-truth labels are single points in the timeline, 
the labels may be very sparse under this sampling strategy. 
To introduce more supervision signals, we define the ground-truth probability distribution following Zhang \textit{et al} \cite{zhang_MATN_2021}.
Specifically, the ground-truth probability at time $t$ is $(y^\xi)_t\in\mathbb{R}$ :
\begin{equation}
    (y^\xi)_t = \textmd{exp}\left( - \frac{t-t_\xi}{2\sigma_\xi^2}  \right ),\ \xi=\{s,m,e\},
\end{equation}
where superscript $s,m,e$ represents start, middle, and end point, 
and $t_s$ and $t_e$ are the ground-truth start and end time, respectively. 
The ground-truth middle time $t_m$ and standard deviation $\sigma_\xi$ are defined as:
\begin{empheq}[left=\empheqlbrace]{align} 
    & t_m=(t_s+t_e)/2, \\
    & \sigma_\xi=\alpha_\xi(t_e-t_s),\ \xi=\{s,m,e\},
\end{empheq}
where $\alpha_{\xi}>0$ is a factor to control the standard deviation.

Our loss function is composed of three parts. For the ordinary network, we apply a weighted binary Cross Entropy Loss (CELoss) \cite{zhang_MATN_2021} during training, given by:
\begin{equation}
    \mathcal{L}_{\textmd{ord}} = \sum_{t=T-M_p+1}^{T} \sum_{\xi}^{s,m,e} 
    (w^\xi)_t
    \textmd{CELoss} \left ((y^\xi)_t, (p^\xi_{\textmd{ord}})_t \right ),
\end{equation}
where $(p^\xi_{\textmd{ord}})_t$ is the estimated probability at time $t$, and $(w^\xi)_t$ is the re-weighting term used for hard sample mining:
\begin{equation}
    (w^\xi)_t = \left | (y^\xi)_t - (p^\xi_{\textmd{ord}})_t \right |^{\gamma},
\end{equation}
where $\gamma$ is the re-weighting factor. 
We can similarly obtain the loss $\mathcal{L}_{\textmd{pro}}$ for the prophet network. 
Afterwards, we employ knowledge distillation to transfer knowledge from the prophet to the ordinary network for future-aware inference:
\begin{equation}
    \mathcal{L}_{\textmd{kd}} = \sum_{t=T-M_p+1}^{T} \sum_{\xi}^{s,m,e} 
    (w^\xi)_t
    \textmd{CELoss} \left ((p^\xi_{\textmd{pro}})_t, (p^\xi_{\textmd{ord}})_t \right ).
\end{equation}
Finally, we derive the total loss $\mathcal{L}$ as follows:
\begin{equation}
    \mathcal{L} = (1-\lambda) \mathcal{L}_{\textmd{ord}} + \mathcal{L}_{\textmd{pro}} + \lambda \mathcal{L}_{\textmd{kd}},
\end{equation}
where $\lambda$ is the trade-off factor.

\subsubsection{Online Inference Implementation} \label{sec:online_inference}
During the inference phase, 
video frames are streamed to the model ceaselessly. 
We maintain a first-in, first-out (FIFO) queue to store the latest $M_h+M_p$ frames, 
which are then fed to the model at each time $T$. 
However, 
this process is inefficient, 
as there is only a single new frame updated at each time. 
To address this issue, 
we propose reusing intermediate results to optimize online inference speed.
Specifically,
we modify the implementation of the first layer of the feature compressor (details provided in the \autoref{sec:apd_online_inference}) to reduce the time complexity 
from $O(M_hnd)$ to an amortized $O(nd)$. 
The runtime analysis is discussed in \autoref{sec:runtime_analysis}.


\section{Experiments}

\begin{table*}[t]

\caption{
Performance comparison with the modified offline baselines and online baselines,
and  the ablation studies on Language-guided Feature Compressor (LFC) and Prophet Decoder (PD). 
``-l'': language branch;
``-v'': vision branch;
``Ho'': historical frames;
``FP'': future prediction.
%
}
\vspace{-1em}

\resizebox{\textwidth}{!}{
\begin{tabular}{ccccccccccccccc}
\hline
\multicolumn{1}{c|}
{\multirow{2}{*}{\textbf{Model}}} & 
\multicolumn{4}{c|}{\textbf{ActivityNet Caption}}& 
\multicolumn{4}{c|}{\textbf{TACoS}}&
\multicolumn{6}{c}{\textbf{MAD}}\\ 
\cline{2-15} 
\multicolumn{1}{c|}{}                       & $R^1_{0.5}$ & $R^1_{0.7}$ & $R^5_{0.5}$ & \multicolumn{1}{c|}{$R^5_{0.7}$} & $R^1_{0.5}$ & $R^1_{0.7}$ & $R^5_{0.5}$ & \multicolumn{1}{c|}{$R^5_{0.7}$} & $R^5_{0.3}$ & $R^5_{0.5}$ & $R^{50}_{0.3}$ & $R^{50}_{0.5}$ & $R^{100}_{0.3}$ & $R^{100}_{0.5}$ 

\\ \hline
\multicolumn{15}{c}{\textit{Comparision with \textbf{modified offline baselines}}} \\                                                              \hline
\multicolumn{1}{c|}{VSLNet \cite{zhang_vslnet_2020}}                 & 12.89       & 5.05       & 25.24            & \multicolumn{1}{c|}{8.79}           & 25.74       & 12.60       & 35.97           & \multicolumn{1}{c|}{16.50}           & -        & -       & -        & -        & -         & -         \\
\multicolumn{1}{c|}{2DTAN \cite{zhang_2dtan_2020}}                  & 8.39       & 2.96       & 28.13       & \multicolumn{1}{c|}{12.82}       & 6.82       & 3.32       & 22.72       & \multicolumn{1}{c|}{9.42}       & -        & -        & -        & -        & -         & -         \\
\multicolumn{1}{c|}{SeqPAN \cite{zhang_seqpan_2021}}                 & 12.57       & 4.79       & 24.83           & \multicolumn{1}{c|}{8.64}           & 25.07       & 13.67       & 37.79           & \multicolumn{1}{c|}{19.52}           & -        & -        & -        & -        & -         & -          \\
\multicolumn{1}{c|}{SMIN \cite{wang_smin_2021}}                   & 7.47       & 2.64       & 24.74       & \multicolumn{1}{c|}{11.28}       & 6.00       & 2.92       & 19.99       & \multicolumn{1}{c|}{8.29}       & -        & -        & -        & -        & -         & -          \\ \hline
 	 	 	 	 	 	 	 
\multicolumn{15}{c}{\textit{Comparision with \textbf{online baselines}}}                                                                                                                                                                                                                                 \\ \hline
\multicolumn{1}{c|}{OadTR \cite{wang_oadtr_2021}}                  & 23.27       & 10.97       & 48.15       & \multicolumn{1}{c|}{29.76}       & 21.12       & 10.92       & 37.99       & \multicolumn{1}{c|}{21.09}       & 2.50        & 0.90        & 8.61         & 4.12         & 13.13         & 6.30          \\
\multicolumn{1}{c|}{LSTR \cite{xu_lstr_2021}}                   & 24.05       & 11.19       & 50.77       & \multicolumn{1}{c|}{31.63}       & 26.02       & 16.75       & 43.01       & \multicolumn{1}{c|}{27.99}       & 3.56        & 1.43        & 11.73        & 4.99         & 17.50         & 7.64          \\
\multicolumn{1}{c|}{GateHUB \cite{chen_gatehub_2022}}                & 23.30       & 11.31       & 50.25       & \multicolumn{1}{c|}{31.00}       & 27.10       & 17.25       & 43.44       & \multicolumn{1}{c|}{26.87}       & 3.38        & 1.47        & 11.96        & 4.74         & 18.20         & 7.56          \\ \hline
\multicolumn{15}{c}{\textit{\textbf{Ablation studies}}} \\ \hline
\multicolumn{1}{c|}{Ours w/o LFC-l}         & 25.18       & 12.48       & 52.42       & \multicolumn{1}{c|}{32.43}       & 28.81       & 18.24       & 47.34       & \multicolumn{1}{c|}{30.27}       & 4.20        & 1.60        & 15.65        & 7.12         & 21.79         & 10.80         \\
\multicolumn{1}{c|}{Ours w/o LFC-v}         & 25.31       & 12.53       & 53.25       & \multicolumn{1}{c|}{33.26}       & 29.43       & 18.76       & 47.62       & \multicolumn{1}{c|}{30.73}       & 4.41        & 1.77        & 15.86        & 7.46         & 22.30         & 11.04         \\
\multicolumn{1}{c|}{Ours w/o LFC}           & 24.90       & 12.43       & 51.60       & \multicolumn{1}{c|}{31.62}       & 28.39       & 17.50       & 46.71       & \multicolumn{1}{c|}{29.49}       & 3.87        & 1.40        & 15.10        & 6.77         & 21.12         & 10.20         \\
\multicolumn{1}{c|}{Ours w/ FP}             & 21.94       & 10.65       & 49.73       & \multicolumn{1}{c|}{31.17}       & 26.60       & 16.71       & 43.76       & \multicolumn{1}{c|}{28.90}       & 3.08        & 1.15        & 13.33        & 5.71         & 19.64         & 8.76          \\
\multicolumn{1}{c|}{Ours w/o HoPD}          & 24.92       & 12.31       & 53.52       & \multicolumn{1}{c|}{33.61}       & 29.44       & 18.83       & 47.93       & \multicolumn{1}{c|}{31.14}       & 4.37        & 1.88        & 15.84        & 7.60         & 22.05         & 10.88         \\
\multicolumn{1}{c|}{Ours w/o PD}            & 23.34       & 11.46       & 52.90       & \multicolumn{1}{c|}{32.81}       & 28.60       & 17.72       & 47.06       & \multicolumn{1}{c|}{30.74}       & 3.21        & 1.20        & 14.33        & 6.14         & 20.46         & 9.41          \\ \hline
\multicolumn{1}{c|}{\textbf{Ours}}                   & \textbf{25.48}       & \textbf{12.56}       & \textbf{53.77}       & \multicolumn{1}{c|}{\textbf{33.70}}       & \textbf{29.74}       & \textbf{19.07}       & \textbf{48.11}       & \multicolumn{1}{c|}{\textbf{31.19}}       & \textbf{4.71}        & \textbf{2.00}        & \textbf{16.34}        & \textbf{7.80}         & \textbf{22.61}         & \textbf{11.45}         \\ \hline
\end{tabular}
}


\label{table:baselines}

\end{table*}




\subsection{Datasets}
There are several video datasets available that contain paired video and sentence query annotations at the temporal level. To conduct our experiments, we selected datasets that are sufficiently long to simulate streaming videos, namely
ActivityNet Captions, TACoS, and MAD datasets,
and constructed the data for our task as described in \autoref{sec:train_infer}. 

\textbf{ActivityNet Captions} \cite{krishna_anetcap_2017}. 
This dataset is introduced by Krishna \textit{et al.} \cite{krishna_anetcap_2017} for dense video captioning. It consists of 14,926 videos and 71,953 annotations. The average length of the video and the average length of the events are 117.6s and 37.14s, respectively. Its vocabulary size is 15,505.

\textbf{TACoS} \cite{tacos_2013}. It derived from the MPII Cooking Composite Activities dataset \cite{regneri2013grounding}, and is composed of 127 videos accompanied by 18,227 annotations. The average video length is 286.59s, while the average event length is 27.88s. The vocabulary size within this dataset is only 228.

\textbf{MAD} \cite{soldan2022mad}. Each video in MAD is a complete movie with a duration of 110 minutes on average \cite{soldan2022mad}. It contains 650 videos and 348,600 annotations.  
Notably, the dataset exhibits a small average event length of 4.1s and possess a vast vocabulary composed of 61,400 unique terms.


\subsection{Experimental Settings}

\subsubsection{Evaluation Metrics}
To evaluate the effectiveness of our TSGSV model,
we conducted an offline evaluation 
(i.e., evaluating after localizing the video clip).
The primary objective of this approach is to assess the accuracy of the localization process.
This approach is the same with prior research on live-stream highlight detection~\cite{zhao_antpivot_2022},
which also employed a similar offline evaluation method even though they proposed an online method. 
The method generally aims to evaluate the extent to which the gap between the video and the accompanying language is bridged \cite{NieQM00B22, xiao2022radar}.
%
%


We first acquired the predicted event clips using the TSGSV model. Concretely, the model will produce the probability of start $\mathbf{s}\in\mathbb{R}^L$ and end $\mathbf{e}\in\mathbb{R}^L$, where $L$ is the length of the complete video. The confidence score between language query and event clip starting from $i-1$ to $j$ is $\textmd{cs}_{ij}$:
\begin{equation}
    \textmd{cs}_{ij} = s_i * e_j, i=1,...,L, j=i, ..., L.
\end{equation}
However, the above method has a significant drawback of producing a large number of candidates which ultimately results in memory overhead, consequently slowing down the video grounding process.
Specifically, the method produces $O(L^2)$ candidate moments, which becomes computationally expensive to handle. 
Therefore, 
we adopted the sparse sampling strategy proposed by Zhang \textit{et al.} \cite{zhang_2dtan_2020} to eliminate the redundant moments and reduces the number of candidates to $O(L\textmd{log}L)$.
By doing so, we can maintain a balance between the robustness of the approach and the computational efficiency of the model.

We then applied the metric ``$\textmd{R}@n,\textmd{IoU}=m$'' ($R^n_m$) to evaluate the model. It is defined as the percentage of language queries that have at least one correct clip in the Top-$n$ predicted event clips. A clip is correct when its IoU with the ground truth clip is larger than $m$.

\subsubsection{Implementation Details}
We performed our experiments with 8 Nvidia V100 GPUs. For ActivityNet and TACoS dataset, we apply BERT \cite{devlin_bert_2019} to extract sentence features, and C3D \cite{tran2015learning}to extract frame features. For MAD dataset, both sentence and frame features are extracted by CLIP \cite{radford2021learning}.
We set $d$ to 1024 for ActivityNet and TACoS, and 512 for MAD. The re-weighting factor $\gamma$ is 3. We set $\alpha_s, \alpha_m, \alpha_e$ to be 0.25, 0.21, and 0.25, respectively \cite{liu2023hs}.
To train our model, we used the AdamW \cite{2019_adamw} optimizer with weight decay 
5e-4, 
linear warm-up, and cosine annealing. Our models were optimized with a batch size of 512 for 15 epochs.


\subsection{Comparison with Baselines}
We compared our method with recent state-of-the-art methods, 
including \textbf{modified offline baselines} from temporal sentence grounding methods (VSLNet\cite{zhang_vslnet_2020}, 2DTAN\cite{zhang_2dtan_2020}, SeqPAN\cite{zhang_seqpan_2021}, and SMIN\cite{wang_smin_2021}), and \textbf{online baselines} from Online Action Detection (OAD) methods (OadTR\cite{wang_oadtr_2021}, LSTR\cite{xu_lstr_2021}, GateHUB\cite{chen_gatehub_2022}).

Since the offline baselines can only handle a fixed duration of video,
we made modifications to these baselines to suit streaming scenario. 
Specifically, we partitioned the video into 50\% overlapping chunks of length $M_h$. 
Subsequently, we fed each chunk into the offline baselines, and merge the outputs using NMS.
%
The OAD methods in online baselines are originally designed for predicting the action class at time $T$ given $\mathcal{I}_T$. 
To use them as baselines, 
we treated the feature vector prior to their final classifier as the video representation
$\mathbf{H}$, 
and fed 
query $\mathbf{q}$ with it to the predictor in our method. It should be noted that reinforcement learning approaches~\cite{he_rwmrl_2019, wang_smrl_2019, wu_tsp_prl_2020, sun_maban_2021} cannot be adapted for streaming video due to their reliance on global context modeling, which requires access to the complete video, or the policy networks fails to converge.

Upon analyzing the experimental results presented in \autoref{table:baselines}, 
several key observations can be made.
Firstly, when compared to all the modified OAD methods, 
our proposed approach achieves the best performance. This phenomenon is consistent across all datasets and metrics, which demonstrates the superiority and robustness of our model.
In contrast,
the modified offline models generally yielded inferior results compared to their offline counterparts. 
It is worth to note that, these models failed to converge in the MAD dataset, hence, 
their performance was not reported in the table.
This is not surprising as offline models are not equipped to deal with challenges related to incomplete information inference and historical information compression.


    
    
    %
    %

\subsection{Ablation Studies}
In this section, we carried out a series of experiments to further analyze the efficacy of our proposed model, as reported in \autoref{table:baselines}.  
Specifically, we investigated the impact of the feature compressor and the prophet-decoder in our model. 
Moreover, our analysis offers a possible explanation and deeper understanding of the underlying mechanisms for the effectiveness of the prophet decoder.


\subsubsection{Ablation of feature compressor}
We compared our model with the following variants: 
\textbf{1) w/o LFC-l}, which remove the language branch in the language-guided feature compression (LFC); 
\textbf{2) w/o LFC-v}, which remove the vision branch; and \textbf{3) w/o LFC}, which remove the whole feature compressor. 

The results of all these variants yielded consistently lower performance compared to the full model.
This highlights the effectiveness of the feature compressor in terms of better information condensation and higher accuracy. 
Additionally, it was observed that the performance of \textbf{w/o LFC-l} is inferior than that of \textbf{w/o LFC-v}, which demonstrates that the language branch plays a greater role in the feature compression process.

\subsubsection{Ablation of prophet decoder}
In order to assess the impact of Prophet Decoder (PD), 
we conducted a series of experiments, incorporating the following three variants: 
\textbf{1) w/ FP}, 
wherein the Prophet network was replaced with future prediction~\cite{xu_TRN_2019, wang_oadtr_2021}, 
\textbf{2) w/o HoPD}, removing the historical features of PD; 
and \textbf{3) w/o PD}, removing the whole prophet network.

The results of \textbf{w/ FP} were found to be consistently lower than those of \textbf{w/o PD}.
This suggests that the ability to predict future actions is not a significant factor in improving the temporal grounding performance for our specific task. 
On the other hand, the results of \textbf{w/o PD} yielded consistently lower results than our proposed full model, 
providing empirical evidence for the importance and effectiveness of the prophet-decoder
It is worth noting that, while the performance of \textbf{w/o HoPD} was only marginally inferior to our full model. 
This indicates that the historical feature can still enhance performance when considered by  the teacher network, even though it has already been integrated into the student network.
This may be attributed to the regularization capacity offered by the historical features of PD.

\begin{figure}[t]
    \centering
    \includegraphics[width=0.85\linewidth]{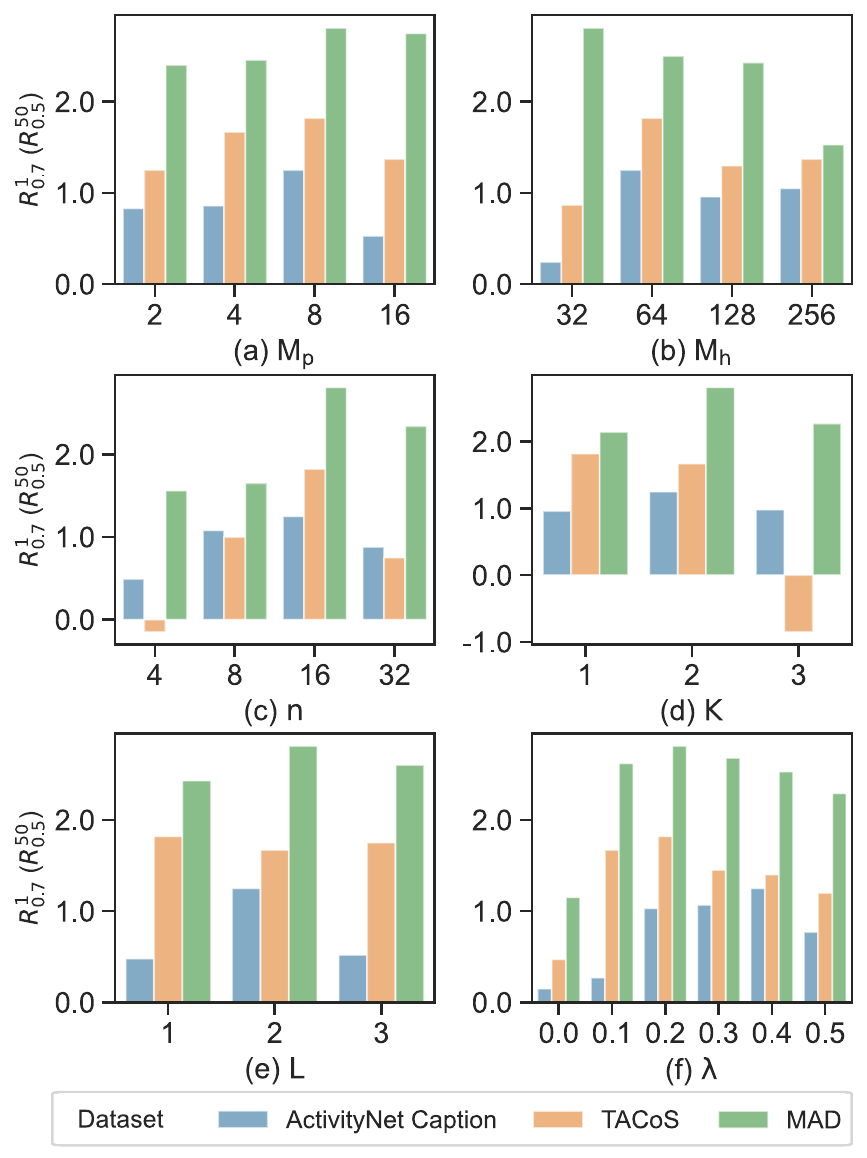}
   \vspace{-1em}
    \caption{Sensitivity analysis of  hyper-parameters. 
    The performance is measured by absolute gain compared with the best online baseline.}
    \label{fig:sensitivity}
\end{figure}

\subsection{Sensitivity Analysis}
We investigated how 
hyper-parameters affect performance, as shown in \autoref{fig:sensitivity}. 
%
Specifically, 
we explored six main parameters: 
present, historical, and compressed feature length 
denoted by 
$M_p$, $M_h$, and $n$, respectively;
the number of feature compressor layers $K$, 
the number of decoder layers $L$, 
and the KD weight $\lambda$. 
%
The results depicted in the figure reveal several important patterns.
First, it was observed that the optimal value of $M_p$ and $n$ were identical across all datasets, indicating that these hyper-parameters may be less dependent on data characteristics.
In contrast, the optimal value of $M_h$ varied across the different datasets, with a value of 64 for ActivityNet and TACoS, and 32 for MAD. 
It is suggested that this discrepancy may be attributable to the smaller average moment length in MAD (4.1s) compared to that in ActivityNet (37.14s) and TACoS (27.88s).
In terms of $K$ and $L$, 
we observed that the optimal values for ActivityNet and MAD were greater than those for TACoS. 
    Notably, the average query length and vocabulary size of ActivityNet and MAD were larger than those of TACoS, implying that the query sentences in TACoS are relatively simple and thus require less model complexity (i.e., smaller $K$ and $L$).
%
    
    
    %
    

\subsection{Runtime Analysis} \label{sec:runtime_analysis}
\begin{table}[t]

\caption{Runtime of our proposed methods with different design choices.}
\vspace{-1em}

\resizebox{\linewidth}{!}{
\begin{tabular}{ccc}
\hline
\multirow{2}{*}{Model}    & \multicolumn{2}{c}{Frames Per Second (FPS)} \\
                          & RGB Feature Extraction      & Our model     \\ \hline
Ours w/o online inference & \multirow{2}{*}{68.52}      & 26.79         \\ \cline{1-1} \cline{3-3} 
Ours                      &                             & 45.81         \\ \hline
\end{tabular}
}
\label{table:runtime}

\end{table}

%

The runtime of our model with various design choices was evaluated using a single 2080Ti GPU and videos from the MAD dataset, with the results presented in \autoref{table:runtime}.
Our inference model included a pre-processing step and an ordinary network. 
The pre-processing step,
mainly comprising video and sentence feature extraction,
operated at a rate of 68.52 FPS. 
Notably, 
the video feature extraction process primarily determined the runtime, 
as the sentence feature computation was performed only once for each query and thus had negligible amortized runtime.
In contrast, the ordinary network without online inference design had a runtime limited to 26.79 FPS, as detailed in \autoref{sec:online_inference}.
Overall, our inference optimization strategy yielded a significant acceleration of our model to enable efficient online inference.

\section{Conclusion and Future Work}
In this paper, 
we work towards temporal sentence grounding in streaming videos. 
We design a Twin-net architecture to compensate for the absence of future frames during online inference. 
Additionally, we developed an innovative language-guided feature compressor that refines historical frames, thereby optimizing the grounding accuracy, while simultaneously reducing computational complexity. 
Through extensive experimental evaluation, we demonstrated the efficacy and superiority of our approach.

In future work, we aim to extend our model for streaming video-text pretraining~\cite{GAN_CVPR_2023}, 
providing benefits for a variety of streaming video understanding tasks. By exploring the potential of our proposed approach, we hope to contribute to the advancement of temporal sentence grounding techniques in real-world scenarios.

\section{Acknowledgements}
This work is supported by the National Natural Science Foundation of China, No.: 62176137, and No.: 62006140; the Shandong Provincial Natural Science and Foundation, No.: ZR2020QF106; Ant Group.

\clearpage



%



\bibliographystyle{IEEEtran}

\bibliography{citations}


%

\balance

\begin{appendices}

\section{Online Inference} \label{sec:apd_online_inference}
In this section, we illustrate how to modify the implementation of the first layer of the Language-guided Feature Compressor (LFC) to reduce its amortized time complexity.

\subsection{Complexity Analysis}
The LFC consists of vision branch, language branch, and fusion network. We only analyze the vision branch, because the language branch shares a similar architecture with the vision branch, and the fusion network has relatively small time complexity compared to them.

The first layer of the vision branch takes historical/future feature sequence $\mathbf{V}^0\in\mathbb{R}^{M_h\times d}$ as input, where $d$ is the embedding dimensions and $M_h$ is the length of historical feature. It calculates the importance logits of each input frame to output tokens 
$\textbf{E}^1_V\in\mathbb{R}^{M_h\times n}$ 
using a Multi-Layer Perceptron (MLP):
\begin{equation} \label{eq:mlp_mat}
    \textbf{E}^1_V = \textmd{MLP}^1_V \left ( \mathbf{V}^0 \right ),
\end{equation}
where MLP is composed of two fully connected layers. There are $M_hdn + M_hn^2$ multiplication operations in total. Considering that $d \gg n$, the time complexity is $O(M_hnd)$.

\subsection{Implementation Optimization}
During inference, the video frames are streamed to the
model ceaselessly. We feed the latest $M_h +M_p$ frames in the FIFO queue into our model at each time T, where $M_p$ is the length of present memory. Because there is only a single new frame updated each time (i.e., frame at time $T$), we can reuse previous intermediate results for optimization. 

For a clearer illustration, we rewrite $\mathbf{V}^0$ in the row vector form:
\begin{equation}
    \begin{bmatrix}
        (\mathbf{v^0_1})^T \\
        (\mathbf{v^0_2})^T \\
        ... \\
        (\mathbf{v^0_t})^T \\
        ... \\
        (\mathbf{v^0_{M_h}})^T
    \end{bmatrix},
\end{equation}
where $\mathbf{v^0_t}\in\mathbb{R}^d$ is the $t$-th frame vector. Thus, the \autoref{eq:mlp_mat} can be re-written as:
\begin{equation}
    \textbf{E}^1_V = \begin{bmatrix}
        (\mathbf{W}_2(\mathbf{W}_1\mathbf{v^0_1}+\mathbf{b}_1) + \mathbf{b}_2)^T \\
        (\mathbf{W}_2(\mathbf{W}_1\mathbf{v^0_2}+\mathbf{b}_1) + \mathbf{b}_2)^T \\
        ... \\
        (\mathbf{W}_2(\mathbf{W}_1\mathbf{v^0_{M_h}}+\mathbf{b}_1) + \mathbf{b}_2)^T
    \end{bmatrix},
\end{equation}
where $\mathbf{W}_1\in\mathbb{R}^{n\times d}, \mathbf{W}_2\in\mathbb{R}^{n\times n}$ are transformation matrices, and $\mathbf{b}_1\in\mathbb{R}^{n}, \mathbf{b}_2\in\mathbb{R}^{n}$ are bias vectors. At each time, we only need to pop up the oldest vector $\mathbf{W}_2(\mathbf{W}_1\mathbf{v^0_1}+\mathbf{b}_1) + \mathbf{b}_2$ and calculate the new vector $\mathbf{W}_2(\mathbf{W}_1\mathbf{v^0_{M_h}}+\mathbf{b}_1) + \mathbf{b}_2$. Thus, the amortized time complexity is reduced from $O(M_hnd)$ to $O(nd)$.

\end{appendices}

\endgroup

\end{document}